\documentclass[11pt]{article}
\usepackage[final]{acl}
\usepackage{times}
\usepackage{url}
\usepackage{latexsym}
\usepackage{booktabs}
\usepackage{graphicx}
\usepackage{enumitem}
\usepackage{array}
\usepackage{tabularx}
\usepackage{multirow} 
\usepackage{float}

\newcommand{\logo}[2]{%
  \raisebox{-0.25\height}{\includegraphics[width=#2]{#1}}}

\title{\textit{What am I missing here?}: Evaluating Large Language Models for Masked Sentence Prediction}

\author{
    Charlie Wyatt\thanks{\ \texttt{charles.wyatt@student.unsw.edu.au}} \quad Aditya Joshi \quad Flora Salim \\
    University of New South Wales, Sydney, Australia \\
}

\date{}

\begin{document}
\maketitle

\begin{abstract}
Transformer-based models primarily rely on Next Token Prediction (NTP), which predicts the next token in a sequence based on the preceding context. However, NTP’s focus on single-token prediction often limits a model’s ability to plan ahead or maintain long-range coherence, raising questions about how well LLMs can predict longer contexts, such as full sentences within structured documents. While NTP encourages local fluency, it provides no explicit incentive to ensure global coherence across sentence boundaries—an essential skill for reconstructive or discursive tasks. To investigate this, we evaluate three commercial LLMs (GPT-4o, Claude 3.5 Sonnet, and Gemini 2.0 Flash) on Masked Sentence Prediction (MSP) — the task of infilling a randomly removed sentence — from three domains: ROCStories (narrative), Recipe1M (procedural), and Wikipedia (expository). We assess both \textbf{fidelity} (similarity to the original sentence) and \textbf{cohesiveness} (fit within the surrounding context). 
Our key finding reveals that commercial LLMs, despite their superlative performance in other tasks, are poor at predicting masked sentences in low-structured domains, highlighting a gap in current model capabilities.
\end{abstract}

\section{Introduction}

Large Language Models (LLMs) have rapidly advanced natural language processing, achieving strong performance across a wide range of benchmarks~\cite{anthropic2024claude35sonnet, openai2024gpt4ocard, google2024gemini2}. These models are trained with Next Token Prediction (NTP), a token-level objective that rewards fluent continuation. However, NTP struggles with tasks that require long-term planning, coherence across extended contexts, or discourse-level structure~\cite{bachmann2024pitfallsnexttokenprediction, maharana2024evaluatinglongtermconversationalmemory}.

This trade-off raises a fundamental question: To what extent do LLMs understand and model sentence-level structure within broader document context? Can they generate missing content that is not only fluent but also faithful and contextually grounded?

We argue that evaluating LLMs solely on token-level fluency masks deeper limitations in their ability to reason over and reconstruct global context. This is particularly relevant for applications like summarization, document editing, or repair, where sentence-level understanding is essential.

To explore these limitations, we study how LLMs handle sentence-level uncertainty using the task of \textbf{Masked Sentence Prediction (MSP)} across fidelity and cohesiveness.

We apply MSP to three distinct domains: narrative (ROCStories), procedural (Recipe1M), and expository (Wikipedia). Across all three, we evaluate generations from GPT-4o, Claude 3.5 Sonnet, and Gemini 2.0 Flash. We find that while fidelity is generally low, human annotators often judge the outputs as contextually appropriate. Notably, fidelity improves in more structured domains like Recipe1M, while perceived cohesion suffers.

Such generations reveal a limitation in document-level understanding—one that poses risks in contexts requiring factual accuracy and precise reconstruction, like legal, historical, or journalistic documents.

\section{Methodology}
\label{sec:methodology}

We define MSP as the task of infilling a missing sentence $s_i$ within a document $D$. For a document $D = \{s_0, \dots, s_m\}$, a model $M$ is asked to generate a sentence $s' = M(D - s_i)$. 

\begin{figure*}[t!]
    \centering
    \includegraphics[width=0.7\linewidth]{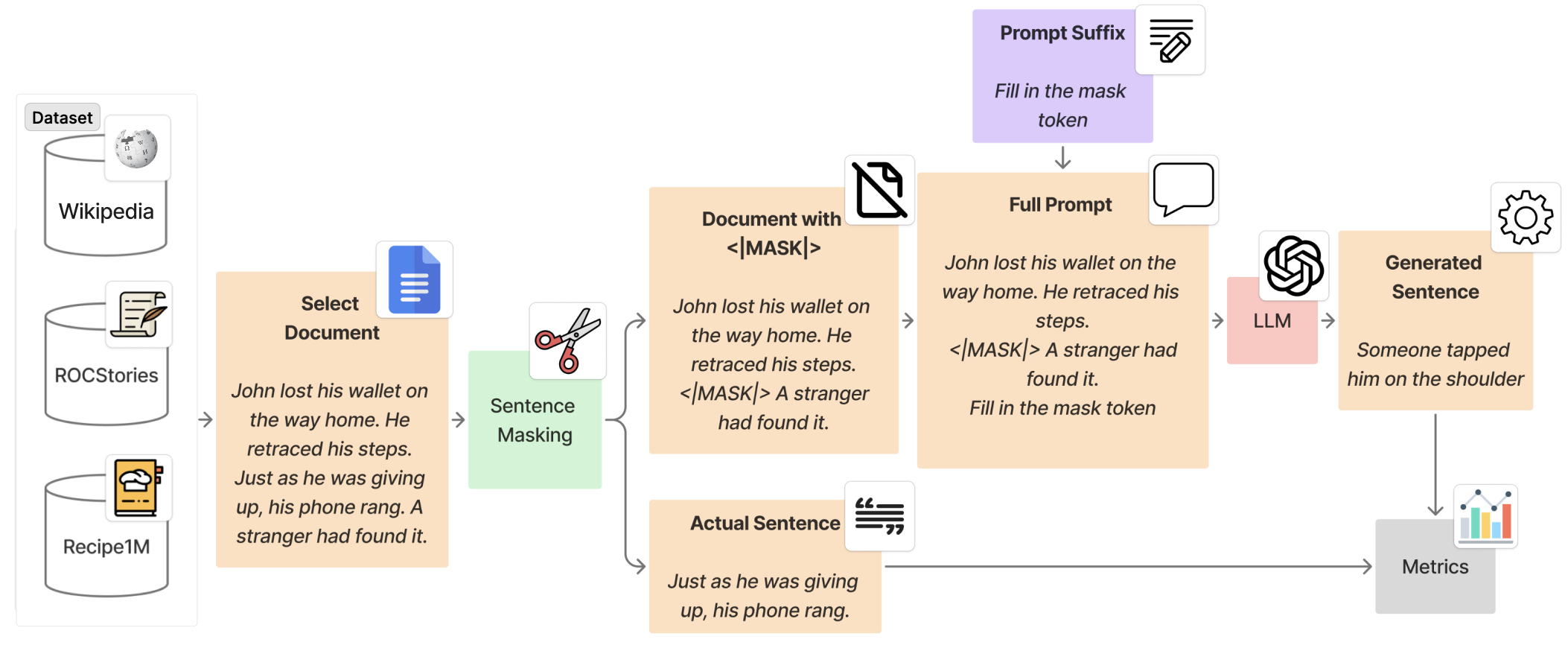}
    \caption{Our experimental pipeline to evaluate masked sentence prediction.}
    \label{fig:methodology}
\end{figure*}

Figure~\ref{fig:methodology} illustrates our MSP approach. First, we perform \textbf{sentence segmentation} through the \verb|en_core_web_sm| model from spaCy~\cite{spacy2}. Next, we apply \textbf{masking} by replacing a single sentence with the special token \texttt{<|mask\_id|>}, following the masking strategies described in Section~\ref{sec:exp_variables}. Finally, for \textbf{sentence generation}, we pass the masked document to the model with the following prompt:

\begin{quote}\small\itshape
\verb|(... text with mask)| Fill in the masked sentence of the text. Do not say anything else. Do not use quotation marks. It should be a complete sentence with punctuation.
\end{quote}

\subsection{Experimental Variables}
\label{sec:exp_variables}
We vary two key conditions to probe model behavior:

\textbf{1. Text Domain:} We evaluate across three domains—narrative (\textsc{ROCStories}), procedural (\textsc{Recipe1M}), and expository (\textsc{Wikipedia}).

\textbf{2. Masking Strategy:} We manipulate context by changing (a) \emph{mask position}, masking the first, last, or a middle sentence, and (b) \emph{mask density}, masking multiple contiguous sentences to assess performance under larger information gaps.

\begin{table*}[t!]
  \centering
  \footnotesize  
  \setlength{\tabcolsep}{4pt} 
  \caption{
    Fidelity scores for Masked Sentence Prediction.
    Full results are in Appendix~\ref{app:detailed_results}.
  }
  \label{tab:fidelity_scores_simple}
  \begin{tabular}{|l|l|*{4}{c|}}
    \hline
    \textbf{Model} & \textbf{Dataset} & \textbf{BLEURT} &
      \textbf{SBERT} & \textbf{ROUGE-1} & \textbf{BLEU} \\
    \hline
    \hspace*{0.1cm}\logo{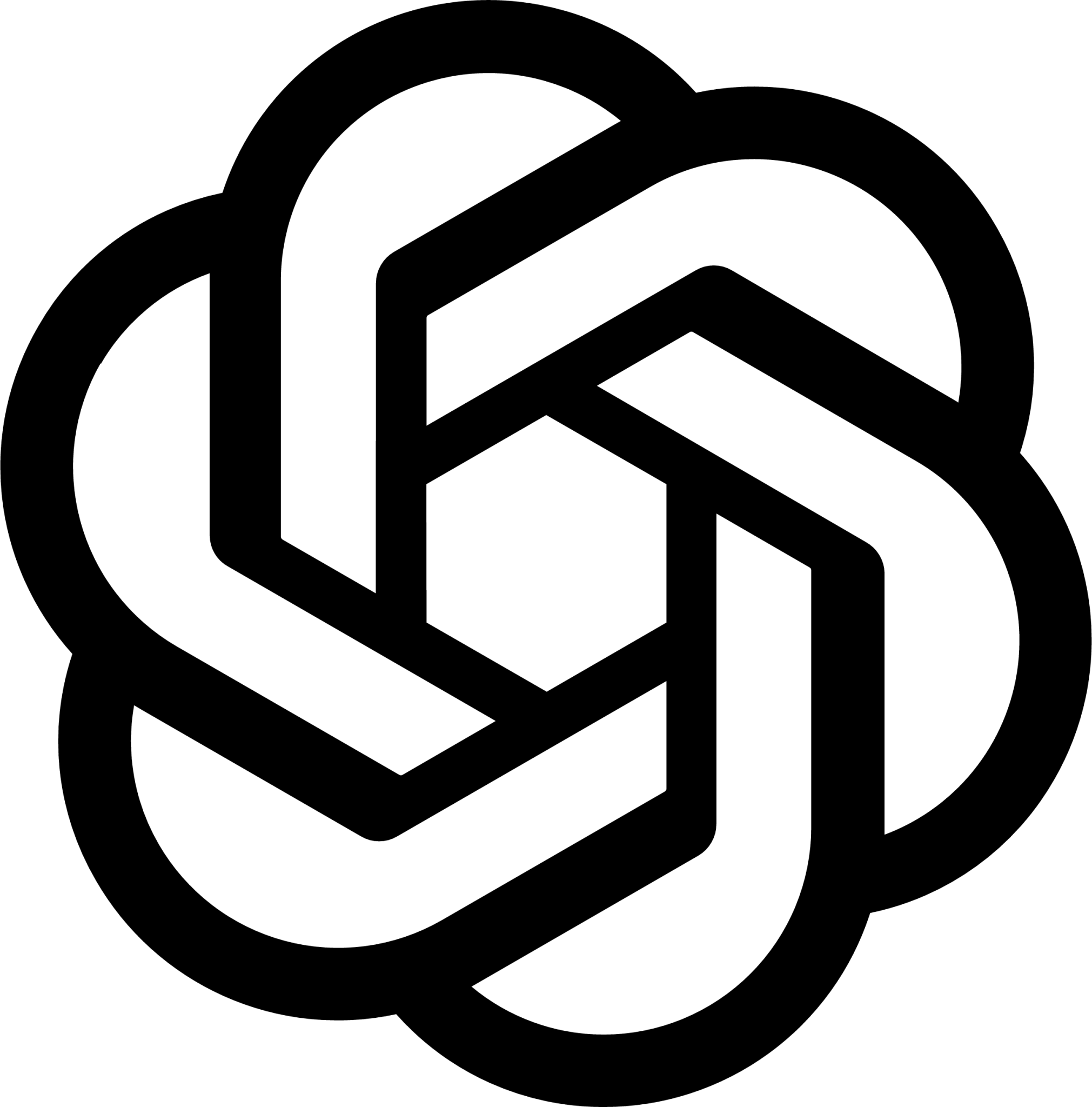}{0.3cm}\; GPT-4o & 
      \textsc{ROCStories} & 0.3839 & 0.4652 & 0.2069 & 0.0225 \\
    \logo{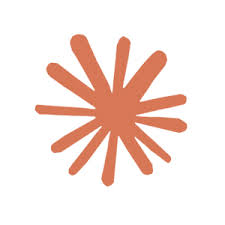}{0.4cm}\; Claude 3.5 Sonnet & 
      \textsc{ROCStories} & \textbf{0.4181} & \textbf{0.5110} & 0.2445 & 0.0272 \\
    \logo{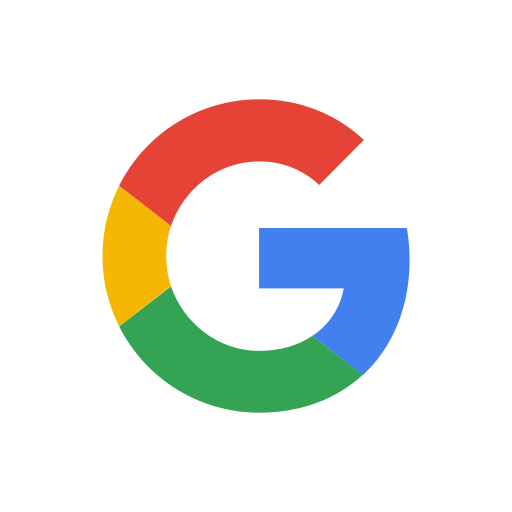}{0.4cm}\; Gemini 2.0 Flash & 
      \textsc{ROCStories} & 0.3747 & 0.4644 & \textbf{0.2455} & \textbf{0.0412} \\
    \hline
    \hspace*{0.1cm}\logo{openai_logo.png}{0.3cm}\; GPT-4o &
      \textsc{Recipe1M} & 0.4987 & 0.5816 & 0.2969 & 0.0913 \\
    \logo{claude_logo.png}{0.4cm}\; Claude 3.5 Sonnet &
      \textsc{Recipe1M} & \textbf{0.5259} & \textbf{0.6082} & \textbf{0.3551} & 0.0980 \\
    \logo{google_logo.png}{0.4cm}\; Gemini 2.0 Flash &
      \textsc{Recipe1M} & 0.4930 & 0.5675 & 0.3261 & \textbf{0.1096} \\
    \hline
    \hspace*{0.1cm}\logo{openai_logo.png}{0.3cm}\; GPT-4o &
      \textsc{Wikipedia} & 0.3248 & 0.4302 & 0.1856 & 0.0257 \\
    \logo{claude_logo.png}{0.4cm}\; Claude 3.5 Sonnet &
      \textsc{Wikipedia} & \textbf{0.3416} & \textbf{0.4396} & \textbf{0.2094} & \textbf{0.0471} \\
    \logo{google_logo.png}{0.4cm}\; Gemini 2.0 Flash &
      \textsc{Wikipedia} & 0.3135 & 0.4064 & 0.1852 & 0.0257 \\
    \hline
  \end{tabular}
\end{table*}

\section{Experimental Setup}
\label{sec:exp-setup}

\subsection{Datasets}
We evaluate across three publicly available corpora:
\begin{itemize}[nosep]
    \item \textbf{Narrative:} \textsc{ROCStories} \citep{mostafazadeh2016corpus}, a set of 5-sentence commonsense narratives.
    \item \textbf{Procedural:} \textsc{Recipe1M} \citep{marin2019recipe1mdatasetlearningcrossmodal}, the cooking–instruction portion of Recipe1M
    \item \textbf{Expository:} \textsc{Wikipedia-2022-English}, encyclopedic articles
\end{itemize}
We randomly sample 400 test documents per dataset to balance statistical power with API constraints.

\subsection{Models}
We evaluate the current flagship LLMs from three major vendors:
\begin{enumerate}[nosep,label=\textbf{\roman*.}]
  \item \textsc{GPT-4o} (OpenAI)  
  \item \textsc{Claude 3.5 Sonnet} (Anthropic)  
  \item \textsc{Gemini 2.0 Flash} (Google)  
\end{enumerate}
All models are accessed via public APIs with default decoding settings (e.g., temperature = 1.0).

\subsection{Evaluation Metrics}
Our evaluation focuses on two behavioral dimensions: \textbf{fidelity}, the similarity between the generated and original sentence, and \textbf{cohesion}, how well the generated sentence fits with the surrounding context.

\subsubsection{Fidelity Metrics (Automatic)}
We use standard similarity metrics to quantify fidelity:
\begin{itemize}[nosep]
    \item \textbf{Semantic:} \textbf{BLEURT}~\citep{sellam-etal-2020-bleurt} and \textbf{SBERT cosine similarity}~\citep{reimers2019sentencebertsentenceembeddingsusing} for semantic similarity.
    \item \textbf{Lexical:} \textbf{ROUGE-1} and \textbf{BLEU}, for n-gram overlap.
\end{itemize}

\subsubsection{Cohesiveness Evaluation (Human)}
To assess cohesion, we conduct a blind human preference test. An annotator was shown both the original and generated sentences (in randomized order) within the document context and asked to indicate which sentence they prefer, if any.
\section{Results}
\label{sec:results}

\subsection{Fidelity}
\label{sec:fidelity}

\label{sec:fidelity_datasets}

As shown in Table~\ref{tab:fidelity_scores_simple}, all models achieve only moderate fidelity, with BLEURT scores rarely exceeding 0.55. This is surprising, given that these corpora were likely seen during pretraining, meaning the original sentence may exist verbatim in the model’s training data.

As shown in Figure 2, Claude 3.5 Sonnet achieves higher average BLEURT scores and lower variance compared to Gemini. This consistency is important for applications like missing value reconstruction in damaged or historical documents, where variations in generated text can cause inaccuracies and compromise content integrity. The wide BLEURT distribution of Gemini shows significant variation between high and low quality generations, potentially limiting its usability

Beyond individual model behavior, we observe that fidelity also strongly correlates with \textbf{domain structure}. Structured domains like \textsc{Recipe1M} consistently yield higher fidelity than the more open-ended \textsc{ROCStories} and \textsc{Wikipedia}. 

\begin{figure}[t!]
\centering
\includegraphics[width=\linewidth]{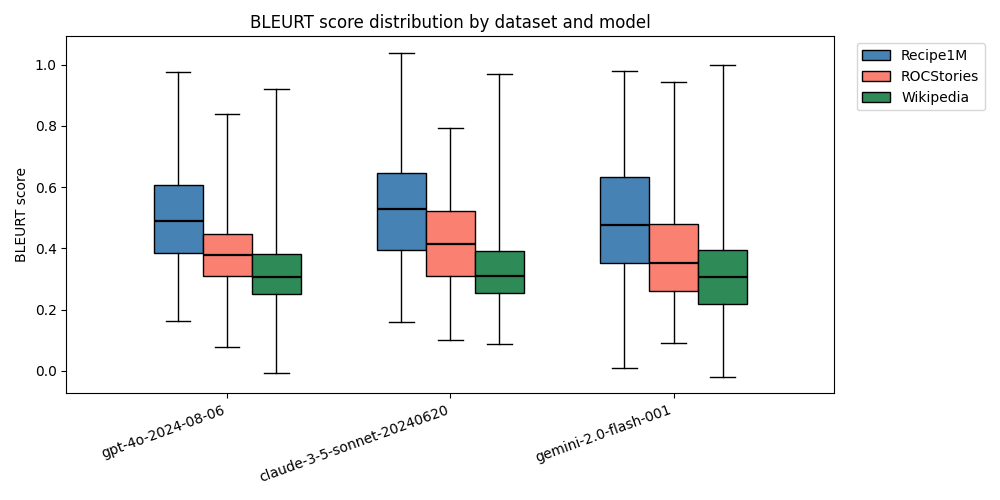}
\caption{
Distribution of BLEURT scores by model and domain.
}
\label{fig:bleurt_boxplot}
\end{figure}

\subsubsection{Qualitative Error Patterns}
\label{sec:fidelity_qual}

Fidelity failures often undermine logical consistency, factual accuracy, or tone. We identified several common failure modes, illustrated with examples in Table~\ref{tab:error_analysis} (Appendix~\ref{app:qualitative_examples}):

\begin{itemize}[leftmargin=*]
\item In \textbf{ROCStories and Wikipedia}, models sometimes generate vague or tonally inconsistent sentences (e.g., inserting overly formal language in casual stories).
\item In \textbf{Recipe1M}, failures often stem from logical or sequencing issues (e.g., placing “Serve immediately” before “Mix ingredients”).
\end{itemize}

These failure modes underscore the importance of separating reconstructive fidelity from contextual cohesiveness: a sentence can be contextually fluent but still pragmatically or factually invalid in structured domains.

\subsubsection{Mask Position Analysis}
\label{sec:fidelity_position}

\begin{figure}
    \centering
    \includegraphics[width=0.8\linewidth]{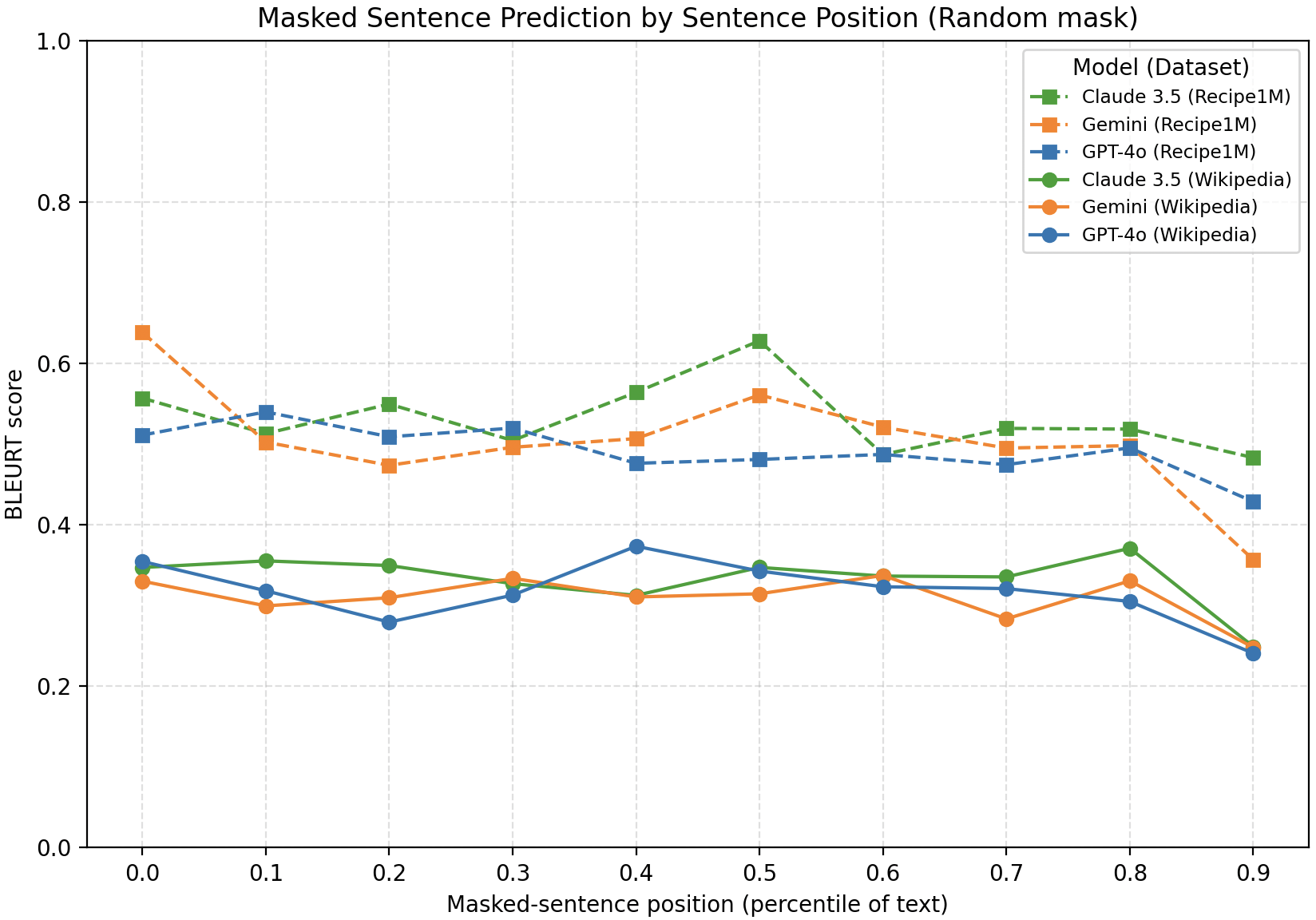}
    \caption{
        Fidelity by sentence position (BLEURT).
    }
    \label{fig:by_position}
\end{figure}

The position of the masked sentence also influences fidelity. As shown in Figure~\ref{fig:by_position}, models perform best when the masked sentence appears in the \textbf{middle} of the text, where both preceding and following context help guide generation.

Masking the \textbf{final} sentence yields the lowest fidelity. This improvement is likely due to the model having access to both preceding and succeeding context, which provides richer information for sentence reconstruction.

\subsubsection{Multi-Sentence Masking}
\label{sec:multi_masking}

\begin{figure}[t!]
    \centering
    \includegraphics[width=0.8\linewidth]{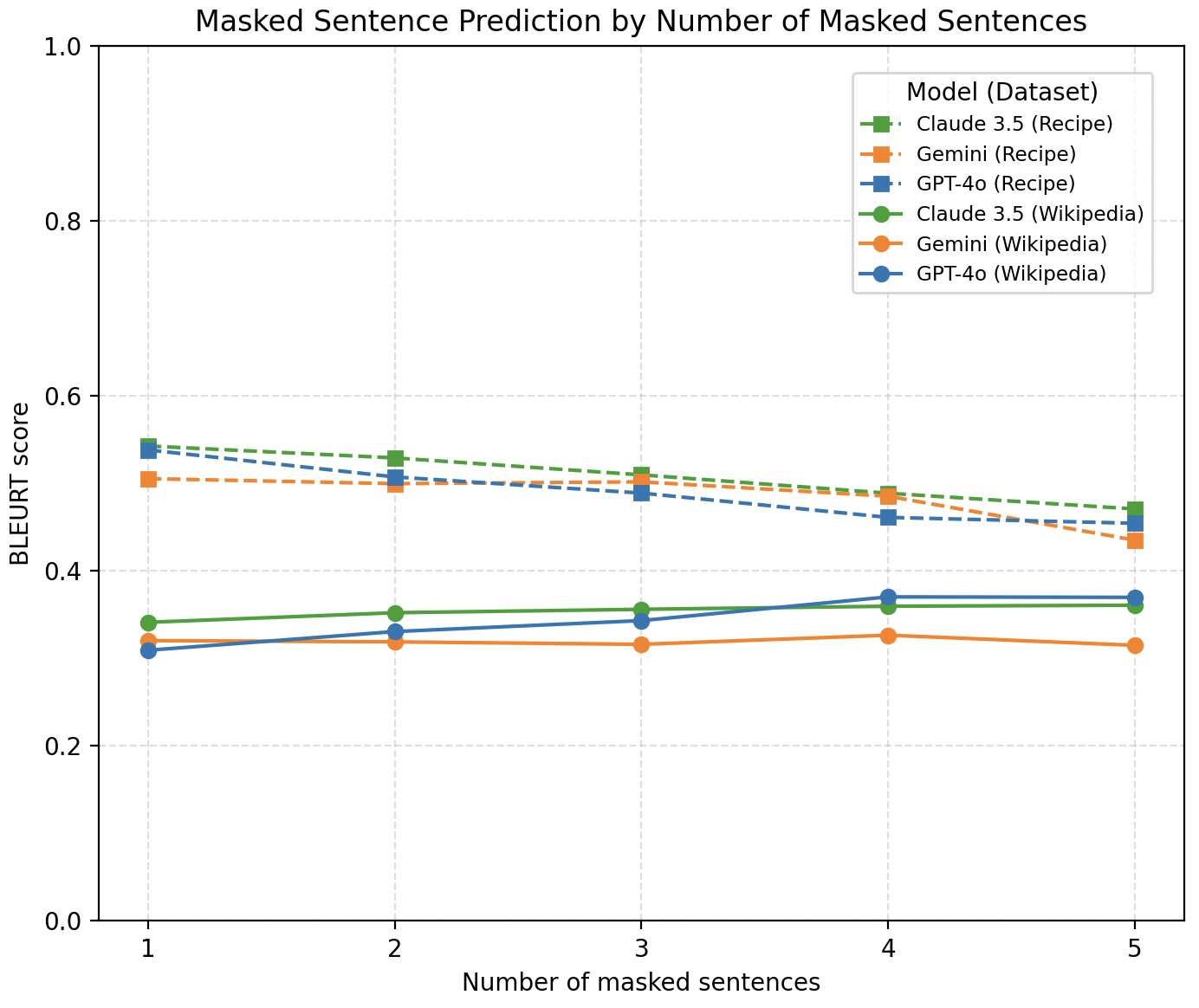}
    \caption{
        \textbf{BLEURT by number of masked sentences.}
    }
    \label{fig:multi_sentence_masking}
\end{figure}

To examine how models handle greater uncertainty, we masked multiple contiguous sentences. Figure~\ref{fig:multi_sentence_masking} shows domain-specific trends:

\begin{itemize}[nosep,leftmargin=*]
    \item In \textsc{Recipe1M}, fidelity declines steadily as more steps are masked— representing the increasing difficulty of the task and the sensitivity of this structured domain to masking.
    \item In \textsc{Wikipedia}, fidelity remains relatively consistent.
\end{itemize}

\begin{table}[t!]
\centering
\small
\setlength\tabcolsep{4pt}
\caption{
    Human preference for generated vs. actual sentences (n=50).
}
\label{tab:human_eval}
\begin{tabular}{llccc}
\toprule
\textbf{Dataset} & \textbf{Preference} & \textbf{GPT-4o} & \textbf{Claude} & \textbf{Gemini} \\
\midrule
\multirow{3}{*}{Stories}
& Equal Preference & 33 & 35 & 32 \\
& Prefer Generated & 8  & 7  & 8  \\
& Prefer Actual    & 9  & 8  & 10 \\
\cmidrule(lr){1-5}
\multirow{3}{*}{Recipes}
& Equal Preference & 28 & 33 & 30 \\
& Prefer Generated & 4  & 8  & 5  \\
& Prefer Actual    & 18 & 9  & 15 \\
\cmidrule(lr){1-5}
\multirow{3}{*}{Wikipedia}
& Equal Preference & 37 & 35 & 38 \\
& Prefer Generated & 3  & 4  & 3  \\
& Prefer Actual    & 10 & 11 & 9  \\
\bottomrule
\end{tabular}
\end{table}

\subsection{Cohesion}
\label{sec:cohesion}

Table~\ref{tab:human_eval} summarizes human preferences between the original and generated sentences. Despite low fidelity scores, generations were often scored with an "Equal Preference".
\textsc{ROCStories} and \textsc{Wikipedia} showed particularly high rates of equal preference (over 60\%), reflecting the models’ ability to generate plausible substitutes in open-ended domains. By contrast, in \textsc{Recipe1M}, annotators preferred the original sentence more often—highlighting the stricter structural demands of procedural text.
Interestingly, this suggests an inverse relationship between fidelity and cohesion: while structured domains like \textsc{Recipe1M} make it easier for models to reproduce the original sentence (higher fidelity), they also make errors more conspicuous to human evaluators. In contrast, the looser expectations in narrative and expository domains allow models to maintain surface-level cohesion even when the generated sentence diverges semantically from the original.

\section{Related Work}
\label{sec:related_work}

MSP is commonly used during pre-training to enhance downstream performance. BART \cite{lewis2019bartdenoisingsequencetosequencepretraining} and T5 \cite{raffel2023exploringlimitstransferlearning} are notable examples, by masking small, contiguous spans of text. These objectives focus on token-level or span-level corruption rather than full-sentence prediction. In contrast, our study treats MSP as a standalone task and evaluates commercial LLMs without fine-tuning.

Beyond pre-training, TIGS \cite{Liu_2019} predicts text by optimizing missing words as parameterized vectors, focusing on shorter text gaps rather than full sentences. INSET \cite{huang-etal-2020-inset} bridges gaps with intermediate sentences using a structured three-step generation process. The Masked Sentence Model (MSM) \citep{zhang2023modelingsequentialsentencerelation} improves cross-lingual dense retrieval by modeling sequential sentence relations using a hierarchical contrastive loss.

While prior work often involves fine-tuning models, we fill a critical gap by assessing commercial LLMs out of the box, highlighting their limitations in predicting masked sentences.

\section{Conclusion \& Future Work}

We evaluated three commercial LLMs—GPT-4o, Claude 3.5 Sonnet, and Gemini 2.0 Flash—on the task of MSP across narrative, procedural, and expository domains. Our results highlight key limitations in current LLMs' ability to model sentence-level coherence and long-range dependencies.

Models perform better in procedurally structured domains like Recipe1M, where the flow is predictable and local context often suffices for accurate reconstruction. In contrast, performance drops in narrative and expository texts, where global coherence and subtle discourse cues are required. This suggests that NTP—which optimizes for short-term fluency—does not adequately support tasks requiring holistic understanding over entire passages.

Future work should explore architectures or training strategies that explicitly capture both local and global context, such as hierarchical attention mechanisms or planning-based generation. Additionally, fine-tuning on MSP-style objectives, incorporating richer prompting strategies (e.g., chain-of-thought), and expanding evaluation to include diverse domains and human judgments could yield deeper insights into model reliability and coherence in real-world settings.

\section*{Limitations}

Our study has several limitations that provide opportunities for future work.

First, our analysis is based on 400 samples from each of the three datasets. While this was sufficient to reveal clear behavioral patterns and provide a meaningful diagnostic signal, larger-scale experiments would be beneficial for ensuring greater statistical significance and for a more fine-grained analysis of rare phenomena.

Second, while we expanded our scope to include narrative, procedural, and expository texts, a broader evaluation across even more diverse genres (e.g., conversational, legal, or poetic text) would be necessary to fully generalize our findings about the interplay between domain structure and generation strategy.

A further limitation is our use of closed-source commercial models. These evaluation datasets were most likely included in the models' vast training corpora. This data contamination could mean that in some instances, the models are performing a form of memorization rather than true, zero-shot generation. Future work could directly address this by replicating our study using a fully open model like OLMo~\citep{groeneveld2024olmoacceleratingsciencelanguage}, whose open training data (e.g., the Dolma corpus) allows for the definitive exclusion of evaluation sets. This would provide a cleaner signal on the models' inherent generative capabilities.

Finally, our human evaluation was conducted by a single annotator who is also an author of this paper. While the evaluations were performed in a blind setting to minimize bias, this setup may still limit the generalizability and objectivity of the results. A larger-scale evaluation with multiple independent annotators would strengthen future conclusions.

\section*{Ethical Considerations}
Since we use publicly available LLMs and datasets there are no known ethical considerations that have been left out.

\paragraph{Data and Model Usage.}
All artifacts used in this study were accessed in accordance with their licenses and terms of service. Our datasets include: \textbf{\textsc{ROCStories}}~\citep{mostafazadeh2016corpus}, distributed under the CC BY 4.0 license; \textbf{\textsc{Recipe1M}}~\citep{marin2019recipe1mdatasetlearningcrossmodal}, used under terms permitting non-commercial research; and articles from \textbf{Wikipedia}, used in accordance with the Creative Commons Attribution-ShareAlike (CC BY-SA) license. All models (\textsc{GPT-4o}, \textsc{Claude 3.5 Sonnet}, and \textsc{Gemini 2.0 Flash}) were accessed via their official public APIs and used in compliance with their respective terms of service. The \texttt{spaCy} library~\citep{spacy2}, used for sentence segmentation, is open-source under the MIT license.

\paragraph{Potential Risks and Societal Impact.}
Our central finding—that LLMs prioritize plausible generation over high-fidelity reconstruction—carries potential societal implications. This behavior is a double-edged sword. On one hand, it can be highly beneficial for creative applications, brainstorming, and generating diverse linguistic paraphrases.

On the other hand, it presents a clear risk. The models' ability to confidently generate plausible but factually incorrect sentences could be misused to create convincing misinformation, to alter the meaning of records, or to automate the production of "authentic-looking" but false content. While our work deals with low-stakes domains like stories and recipes, the underlying behavior we identify is domain-general. We believe it is crucial for the research community and practitioners to be aware of this dual-use nature when deploying such models in high-stakes applications where factual precision is paramount.

\paragraph{Use of AI in Research Preparation.}
We utilized AI-assisted tools during the development of this work. Specifically, ChatGPT was used for drafting, grammatical editing, and proofreading the manuscript. Additionally, GitHub Copilot was employed to assist with coding tasks and debugging. We ensured that all outputs generated by these tools were carefully reviewed and validated by the authors to maintain accuracy and correctness.

\section*{Acknowledgments}
Some of the illustrations in this paper incorporate icons sourced from Flaticon.com. We gratefully acknowledge the individual artists who designed and shared these assets.

\bibliography{acl}

\begin{thebibliography}{16}
\providecommand{\natexlab}[1]{#1}

\bibitem[{Anthropic(2024)}]{anthropic2024claude35sonnet}
Anthropic. 2024.
\newblock \href {https://www.anthropic.com/news/claude-3-5-sonnet} {Claude 3.5 sonnet model card addendum}.
\newblock Accessed: 2025-05-19.

\bibitem[{Bachmann and Nagarajan(2024)}]{bachmann2024pitfallsnexttokenprediction}
Gregor Bachmann and Vaishnavh Nagarajan. 2024.
\newblock \href {https://arxiv.org/abs/2403.06963} {The pitfalls of next-token prediction}.
\newblock \emph{Preprint}, arXiv:2403.06963.

\bibitem[{DeepMind(2024)}]{google2024gemini2}
Google DeepMind. 2024.
\newblock \href {https://blog.google/technology/google-deepmind/google-gemini-ai-update-december-2024/} {Introducing gemini 2.0: our new ai model for the agentic era}.
\newblock Accessed: 2025-05-19.

\bibitem[{Groeneveld et~al.(2024)Groeneveld, Beltagy, Walsh, Bhagia, Kinney, Tafjord, Jha, Ivison, Magnusson, Wang, Arora, Atkinson, Authur, Chandu, Cohan, Dumas, Elazar, Gu, Hessel, Khot, Merrill, Morrison, Muennighoff, Naik, Nam, Peters, Pyatkin, Ravichander, Schwenk, Shah, Smith, Strubell, Subramani, Wortsman, Dasigi, Lambert, Richardson, Zettlemoyer, Dodge, Lo, Soldaini, Smith, and Hajishirzi}]{groeneveld2024olmoacceleratingsciencelanguage}
Dirk Groeneveld, Iz~Beltagy, Pete Walsh, Akshita Bhagia, Rodney Kinney, Oyvind Tafjord, Ananya~Harsh Jha, Hamish Ivison, Ian Magnusson, Yizhong Wang, Shane Arora, David Atkinson, Russell Authur, Khyathi~Raghavi Chandu, Arman Cohan, Jennifer Dumas, Yanai Elazar, Yuling Gu, Jack Hessel, Tushar Khot, William Merrill, Jacob Morrison, Niklas Muennighoff, Aakanksha Naik, Crystal Nam, Matthew~E. Peters, Valentina Pyatkin, Abhilasha Ravichander, Dustin Schwenk, Saurabh Shah, Will Smith, Emma Strubell, Nishant Subramani, Mitchell Wortsman, Pradeep Dasigi, Nathan Lambert, Kyle Richardson, Luke Zettlemoyer, Jesse Dodge, Kyle Lo, Luca Soldaini, Noah~A. Smith, and Hannaneh Hajishirzi. 2024.
\newblock \href {https://arxiv.org/abs/2402.00838} {Olmo: Accelerating the science of language models}.
\newblock \emph{Preprint}, arXiv:2402.00838.

\bibitem[{Honnibal and Montani(2017)}]{spacy2}
Matthew Honnibal and Ines Montani. 2017.
\newblock {spaCy 2}: Natural language understanding with {B}loom embeddings, convolutional neural networks and incremental parsing.
\newblock To appear.

\bibitem[{Huang et~al.(2020)Huang, Zhang, Elachqar, and Cheng}]{huang-etal-2020-inset}
Yichen Huang, Yizhe Zhang, Oussama Elachqar, and Yu~Cheng. 2020.
\newblock \href {https://doi.org/10.18653/v1/2020.acl-main.226} {{INSET}: Sentence infilling with {IN}ter-{SE}ntential transformer}.
\newblock In \emph{Proceedings of the 58th Annual Meeting of the Association for Computational Linguistics}, pages 2502--2515, Online. Association for Computational Linguistics.

\bibitem[{Lewis et~al.(2019)Lewis, Liu, Goyal, Ghazvininejad, Mohamed, Levy, Stoyanov, and Zettlemoyer}]{lewis2019bartdenoisingsequencetosequencepretraining}
Mike Lewis, Yinhan Liu, Naman Goyal, Marjan Ghazvininejad, Abdelrahman Mohamed, Omer Levy, Ves Stoyanov, and Luke Zettlemoyer. 2019.
\newblock \href {https://arxiv.org/abs/1910.13461} {Bart: Denoising sequence-to-sequence pre-training for natural language generation, translation, and comprehension}.
\newblock \emph{Preprint}, arXiv:1910.13461.

\bibitem[{Liu et~al.(2019)Liu, Fu, Liu, and Lv}]{Liu_2019}
Dayiheng Liu, Jie Fu, Pengfei Liu, and Jiancheng Lv. 2019.
\newblock \href {https://doi.org/10.18653/v1/p19-1406} {Tigs: An inference algorithm for text infilling with gradient search}.
\newblock In \emph{Proceedings of the 57th Annual Meeting of the Association for Computational Linguistics}, page 4146–4156. Association for Computational Linguistics.

\bibitem[{Maharana et~al.(2024)Maharana, Lee, Tulyakov, Bansal, Barbieri, and Fang}]{maharana2024evaluatinglongtermconversationalmemory}
Adyasha Maharana, Dong-Ho Lee, Sergey Tulyakov, Mohit Bansal, Francesco Barbieri, and Yuwei Fang. 2024.
\newblock \href {https://arxiv.org/abs/2402.17753} {Evaluating very long-term conversational memory of llm agents}.
\newblock \emph{Preprint}, arXiv:2402.17753.

\bibitem[{Marin et~al.(2019)Marin, Biswas, Ofli, Hynes, Salvador, Aytar, Weber, and Torralba}]{marin2019recipe1mdatasetlearningcrossmodal}
Javier Marin, Aritro Biswas, Ferda Ofli, Nicholas Hynes, Amaia Salvador, Yusuf Aytar, Ingmar Weber, and Antonio Torralba. 2019.
\newblock \href {https://arxiv.org/abs/1810.06553} {Recipe1m+: A dataset for learning cross-modal embeddings for cooking recipes and food images}.
\newblock \emph{Preprint}, arXiv:1810.06553.

\bibitem[{Mostafazadeh et~al.(2016)Mostafazadeh, Chambers, He, Parikh, Batra, Vanderwende, Kohli, and Allen}]{mostafazadeh2016corpus}
Nasrin Mostafazadeh, Nathanael Chambers, Xiaodong He, Devi Parikh, Dhruv Batra, Lucy Vanderwende, Pushmeet Kohli, and James Allen. 2016.
\newblock A corpus and evaluation framework for deeper understanding of commonsense stories.
\newblock \emph{arXiv preprint arXiv:1604.01696}.

\bibitem[{OpenAI et~al.(2024)OpenAI, Hurst, and et~al.}]{openai2024gpt4ocard}
OpenAI, Aaron Hurst, and et~al. 2024.
\newblock \href {https://arxiv.org/abs/2410.21276} {Gpt-4o system card}.
\newblock \emph{Preprint}, arXiv:2410.21276.

\bibitem[{Raffel et~al.(2023)Raffel, Shazeer, Roberts, Lee, Narang, Matena, Zhou, Li, and Liu}]{raffel2023exploringlimitstransferlearning}
Colin Raffel, Noam Shazeer, Adam Roberts, Katherine Lee, Sharan Narang, Michael Matena, Yanqi Zhou, Wei Li, and Peter~J. Liu. 2023.
\newblock \href {https://arxiv.org/abs/1910.10683} {Exploring the limits of transfer learning with a unified text-to-text transformer}.
\newblock \emph{Preprint}, arXiv:1910.10683.

\bibitem[{Reimers and Gurevych(2019)}]{reimers2019sentencebertsentenceembeddingsusing}
Nils Reimers and Iryna Gurevych. 2019.
\newblock \href {https://arxiv.org/abs/1908.10084} {Sentence-bert: Sentence embeddings using siamese bert-networks}.
\newblock \emph{Preprint}, arXiv:1908.10084.

\bibitem[{Sellam et~al.(2020)Sellam, Das, and Parikh}]{sellam-etal-2020-bleurt}
Thibault Sellam, Dipanjan Das, and Ankur Parikh. 2020.
\newblock \href {https://doi.org/10.18653/v1/2020.acl-main.704} {{BLEURT}: Learning robust metrics for text generation}.
\newblock In \emph{Proceedings of the 58th Annual Meeting of the Association for Computational Linguistics}, pages 7881--7892, Online. Association for Computational Linguistics.

\bibitem[{Zhang et~al.(2023)Zhang, Liang, Gong, Jiang, and Duan}]{zhang2023modelingsequentialsentencerelation}
Shunyu Zhang, Yaobo Liang, Ming Gong, Daxin Jiang, and Nan Duan. 2023.
\newblock \href {https://arxiv.org/abs/2302.01626} {Modeling sequential sentence relation to improve cross-lingual dense retrieval}.
\newblock \emph{Preprint}, arXiv:2302.01626.

\end{thebibliography}

\clearpage 
\appendix

\section{Detailed Experimental Results}
\label{app:detailed_results}

This section provides detailed numerical results supplementing the analysis in the main paper. \textbf{Table~\ref{app:tab_full_results}} expands on the main results table by breaking down performance by the position of the masked sentence. \textbf{Tables~\ref{app:tab_recipes_by_length} and \ref{app:tab_recipes_by_proportion}} provide further analysis on the \textsc{Recipe1M} dataset, showing how fidelity scores vary with the total number of sentences and the relative masked position, respectively.

\begin{table*}[!htbp]
    \centering
    \small
    \caption{Full performance breakdown by model, dataset, and masked sentence position. For ROCStories and Recipe1M, "Random" refers to the average score across all positions, while "First" and "Last" refer to masking only the first or last sentence. For Wikipedia, only random middle sentences were masked.}
    \label{app:tab_full_results}
    \begin{tabular}{ll c cccc}
        \toprule
        \textbf{Model} & \textbf{Dataset} & \textbf{Mask Position} & \textbf{BLEURT} & \textbf{SBERT} & \textbf{ROUGE-1} & \textbf{BLEU} \\
        \midrule
        \multirow{3}{*}{GPT-4o} & \multirow{3}{*}{ROCStories} & Random & 0.3839 & 0.4652 & 0.2069 & 0.0225 \\
                                & & First  & 0.3754 & 0.4305 & 0.1807 & 0.0160 \\
                                & & Last   & 0.3754 & 0.4305 & 0.1807 & 0.0160 \\
        \cmidrule(lr){2-7}
        \multirow{3}{*}{Claude 3.5 Sonnet} & \multirow{3}{*}{ROCStories} & Random & 0.4181 & 0.5110 & 0.2445 & 0.0272 \\
                                & & First  & 0.4606 & 0.5803 & 0.2892 & 0.0488 \\
                                & & Last   & 0.4056 & 0.4836 & 0.2076 & 0.0215 \\
        \cmidrule(lr){2-7}
        \multirow{3}{*}{Gemini 2.0 Flash} & \multirow{3}{*}{ROCStories} & Random & 0.3747 & 0.4644 & 0.2455 & 0.0412 \\
                                & & First  & 0.4195 & 0.5468 & 0.2899 & 0.0719 \\
                                & & Last   & 0.3550 & 0.4144 & 0.1931 & 0.0247 \\
        \midrule
        \multirow{3}{*}{GPT-4o} & \multirow{3}{*}{Recipe1M} & Random & 0.4987 & 0.5816 & 0.2969 & 0.0913 \\
                                & & First  & 0.5195 & 0.6224 & 0.3340 & 0.1013 \\
                                & & Last   & 0.4848 & 0.5027 & 0.2439 & 0.0554 \\
        \cmidrule(lr){2-7}
        \multirow{3}{*}{Claude 3.5 Sonnet} & \multirow{3}{*}{Recipe1M} & Random & 0.5259 & 0.6082 & 0.3551 & 0.0980 \\
                                & & First  & 0.5335 & 0.6459 & 0.3811 & 0.1110 \\
                                & & Last   & 0.4858 & 0.5057 & 0.2721 & 0.0705 \\
        \cmidrule(lr){2-7}
        \multirow{3}{*}{Gemini 2.0 Flash} & \multirow{3}{*}{Recipe1M} & Random & 0.4930 & 0.5675 & 0.3261 & 0.1096 \\
                                & & First  & 0.4841 & 0.5897 & 0.3316 & 0.1167 \\
                                & & Last   & 0.4433 & 0.4597 & 0.2255 & 0.0743 \\
        \midrule
        GPT-4o & Wikipedia & Random & 0.3248 & 0.4302 & 0.1856 & 0.0257 \\
        Claude 3.5 Sonnet & Wikipedia & Random & 0.3416 & 0.4396 & 0.2094 & 0.0471 \\
        Gemini 2.0 Flash & Wikipedia & Random & 0.3135 & 0.4064 & 0.1852 & 0.0257 \\
        \bottomrule
    \end{tabular}
\end{table*}

\begin{table*}[H]
\centering
\footnotesize
\caption{BLEURT scores on the \textsc{Recipe1M} dataset, broken down by the total number of sentences in the recipe.}
\label{tab:bleurt_recipe1m_by_length}
\resizebox{\textwidth}{!}{%
\begin{tabular}{l ccccccccccccccc}
    \toprule
    \textbf{Model} & \multicolumn{15}{c}{\textbf{Total Number of Sentences in Recipe}} \\
    \cmidrule(lr){2-16}
    & \textbf{1} & \textbf{2} & \textbf{3} & \textbf{4} & \textbf{5} & \textbf{6} & \textbf{7} & \textbf{8} & \textbf{9} & \textbf{10} & \textbf{11} & \textbf{12} & \textbf{13} & \textbf{14} & \textbf{$\geq$15} \\
    \midrule
    GPT-4o & 0.4939 & 0.4402 & 0.4808 & 0.5149 & 0.4496 & 0.4787 & 0.4670 & 0.5108 & 0.5794 & 0.5169 & 0.5111 & 0.4508 & 0.5551 & 0.4586 & 0.5032 \\
    Claude 3.5 Sonnet & 0.6016 & 0.4333 & 0.4626 & 0.5154 & 0.5039 & 0.5086 & 0.5655 & 0.4804 & 0.5704 & 0.5815 & 0.5490 & 0.5615 & 0.5068 & 0.4726 & 0.5353 \\
    Gemini 2.0 Flash & 0.4105 & 0.3570 & 0.4375 & 0.4800 & 0.4580 & 0.5002 & 0.5433 & 0.4911 & 0.4899 & 0.5578 & 0.4618 & 0.4440 & 0.5507 & 0.5063 & 0.5184 \\
    \bottomrule
\end{tabular}
}
\end{table*}

\begin{table*}[t!]
\centering
\footnotesize
\caption{BLEURT scores on the \textsc{Wikipedia} and \textsc{Recipe1M} datasets, broken down by masked position (0.0–0.9). All experiments used a randomly masked sentence.}
\label{tab:bleurt_by_dataset_and_mask}
\resizebox{\textwidth}{!}{%
\begin{tabular}{llcccccccccc}
\toprule
\textbf{Dataset} & \textbf{Model} & \textbf{0.0} & \textbf{0.1} & \textbf{0.2} & \textbf{0.3} & \textbf{0.4} & \textbf{0.5} & \textbf{0.6} & \textbf{0.7} & \textbf{0.8} & \textbf{0.9} \\
\midrule
Wikipedia & GPT-4o & 0.3546 & 0.3183 & 0.2794 & 0.3129 & 0.3734 & 0.3427 & 0.3232 & 0.3210 & 0.3050 & 0.2408 \\
          & Claude 3.5 Sonnet & 0.3471 & 0.3553 & 0.3495 & 0.3271 & 0.3126 & 0.3472 & 0.3365 & 0.3354 & 0.3709 & 0.2492 \\
          & Gemini 2.0 Flash & 0.3302 & 0.2996 & 0.3096 & 0.3337 & 0.3106 & 0.3145 & 0.3373 & 0.2834 & 0.3306 & 0.2481 \\
\midrule
Recipe1M  & GPT-4o & 0.5111 & 0.5398 & 0.5091 & 0.5198 & 0.4762 & 0.4809 & 0.4872 & 0.4745 & 0.4953 & 0.4287 \\
          & Claude 3.5 Sonnet & 0.5569 & 0.5130 & 0.5496 & 0.5044 & 0.5641 & 0.6280 & 0.4871 & 0.5194 & 0.5185 & 0.4832 \\
          & Gemini 2.0 Flash & 0.6384 & 0.5024 & 0.4737 & 0.4961 & 0.5068 & 0.5609 & 0.5210 & 0.4951 & 0.4981 & 0.3568 \\
\bottomrule
\end{tabular}
}
\end{table*}

\begin{figure}[t!]
    \centering
    \includegraphics[width=\linewidth]{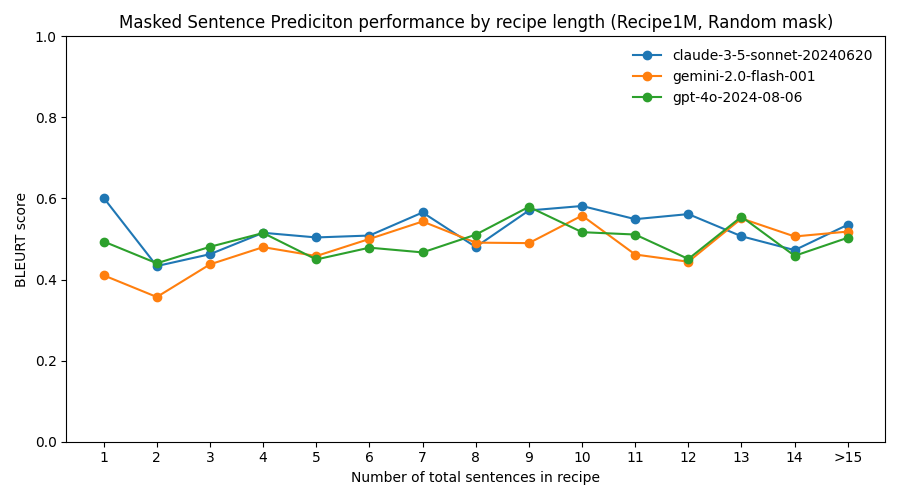}
    \caption{MSP performance (BLEURT) by the number of sentences in the \textsc{Recipe1M} dataset.}
    \label{fig:length_effect}
\end{figure}

\section{Qualitative Generation Examples}
\label{app:qualitative_examples}

This section provides the qualitative examples referenced in Section~\ref{sec:fidelity_qual} of the main text. Table~\ref{tab:qualitative_examples} illustrates the fidelity-plausibility trade-off, while Table~\ref{tab:error_analysis} categorizes common failure modes.

\begin{table*}[t!]
\centering
\small
\caption{
    \textbf{Examples of common failure modes in plausible generation.}
    These sentences are often locally coherent but fail to respect the global context, leading to logical inconsistencies, unresolved narratives, or formatting errors.
}
\label{tab:error_analysis}
\begin{tabularx}{\textwidth}{@{} l >{\raggedright}X >{\raggedright}X l l @{}}
\toprule
\textbf{Model} & \textbf{Generated Sentence} & \textbf{Actual Sentence} & \textbf{Failure Type} & \textbf{Dataset} \\
\midrule
Gemini 2.0 Flash & Keep it in the fridge. & Serve and enjoy. & Logical Inconsistency & Recipe \\
\addlinespace
GPT-4o & In a large bowl, mix together the flour, baking powder, and salt. & 1.In a large bowl, mix together flour, salt, baking powder, cinnamon, nutmeg, ground cloves, ginger and Flax Seeds; set aside. & Formatting Inconsistency & Recipe \\
\addlinespace
Claude 3.5 Sonnet & Frustrated, Ulrich decided to rely on the video's visual demonstrations and hoped he could figure out the repair process without understanding the narration. & Ulrich was able to fix the scanner just by the video alone. & Verbosity & Stories \\
\addlinespace
Gemini 2.0 Flash & Don was shocked, but the CEO insisted it was necessary for the company's future. & Don refused. & Unresolved & Stories \\
\addlinespace
Gemini 2.0 Flash & I had built them a wonderful cat condo. & I had noticed that my cats seemed kind of bored. & No Introduction & Stories \\
\addlinespace
GPT-4o & He asked his wife to pour vinegar over the sting to neutralize the venom. & He scraped the sting with a credit card to help stop the stinging. & More Formal & Stories \\
\bottomrule
\end{tabularx}
\end{table*}

\begin{table*}[t!]
\centering
\small
\caption{
    \textbf{Qualitative examples of the Fidelity-Plausibility Trade-off.}
    High-scoring examples show high fidelity (reconstruction). Low-scoring examples often exhibit high plausibility but low fidelity, where the model generates a coherent but novel sentence (e.g., explaining a dry pond with a "drought" instead of the author's "surprise").
}
\label{tab:qualitative_examples}
\begin{tabular}{@{}llp{3.2cm}p{3.2cm}p{3.2cm}@{}}
\toprule
\textbf{Model} & \textbf{Score Type} & \textbf{Context (with mask)} & \textbf{Actual Sentence} & \textbf{Generated Sentence} \\
\midrule

\textbf{GPT-4o} & High Fidelity & Tom had never been on an airplane. \texttt{<|mask\_id|>} His first time was a bumpy ride. They hit a lot of turbulence. Tom was frightened the whole time. & Tom had never been on an airplane. & Tom had never been on an airplane before. \\
\cmidrule(lr){2-5}
& Low Fidelity & I used to run around a pond. It was there one fall. The next fall it had vanished. There was nothing but dirt. \texttt{<|mask\_id|>} & I was surprised. & A long drought had caused the water to evaporate completely. \\
\midrule

\textbf{Claude 3.5 Sonnet} & High Fidelity & \texttt{<|mask\_id|>} When it came in he added all his favorite apps. He bought a new case. Then as he was out one day, he dropped it face down. Josh shattered his phone within hours of getting it. & Josh was so excited about his new phone. & Josh was excited to receive his brand new smartphone. \\
\cmidrule(lr){2-5}
& Low Fidelity & Tom went to a new church. It was full and a lot of fun. He started going more often. So did a lot of other people. \texttt{<|mask\_id|>} & Eventually Tom wasn't able to fit anymore. & The church grew rapidly as word spread about its welcoming atmosphere and engaging services. \\
\midrule

\textbf{Gemini 2.0 Flash} & High Fidelity & \texttt{<|mask\_id|>} He decides to snack on some peanuts. He eats a handful of peanuts. He is glad that he ate some peanuts. He feels much less hungry afterwards. & John is feeling a little hungry. & John is feeling a little hungry. \\
\cmidrule(lr){2-5}
& Low Fidelity & Before led bike lights, I used battery powered lights. The batteries wore out too quickly. \texttt{<|mask\_id|>} I tried a generator, but it never produced much light. I found an incandescent flashing light which used less energy. & I drove forty five minutes in the dark. & Replacing them was expensive and inconvenient. \\
\bottomrule
\end{tabular}
\end{table*}
\end{document}